%% file: main.tex
\documentclass[10pt,twocolumn,letterpaper]{article}

\usepackage[accsupp]{axessibility}
\usepackage{cvpr}              

\usepackage{graphicx}
\usepackage{amsmath}
\usepackage{amssymb}
\usepackage{booktabs}
\usepackage{multirow}
\usepackage{rotating}
\usepackage{xcolor}

\usepackage[pagebackref,breaklinks,colorlinks]{hyperref}

\definecolor{myPurple}{rgb}{0.4, .0, .8}
\definecolor{myGreen}{rgb}{0, .8, .3}
\definecolor{myRed}{rgb}{0.8, .2, .2}
\definecolor{myOrange}{rgb}{0.7, 0.45, 0.2}
\definecolor{myBlue}{rgb}{.0, .0, 1.0}
\definecolor{myBlue2}{rgb}{.0, .0, 0.5}
\definecolor{myBlack}{rgb}{.0, .0, 0.0}

\newcommand{\ve}[1]{\mbox{{\bf #1}}} 

\newcommand{\mysubsubsection}[1]{\TODO{use paragraph!}}

\usepackage[capitalize]{cleveref}
\crefname{section}{Sec.}{Secs.}
\Crefname{section}{Section}{Sections}
\Crefname{table}{Table}{Tables}
\crefname{table}{Tab.}{Tabs.}

\def\cvprPaperID{33} 
\def\confName{CVPR}
\def\confYear{2023}

\input{preamble.tex}
\begin{document}

\title{NeuMap: Neural Coordinate Mapping by \\ Auto-Transdecoder for Camera Localization}


\author{Shitao Tang \quad {Sicong Tang} \quad {Andrea Tagliasacchi} \quad {Ping Tan} \quad {Yasutaka Furukawa} \\
	Simon Fraser University \\
	{\tt\small \{shitao\_tang, sta105, andrea.tagliasacchi, pingtan, furukawa\}@sfu.ca}
}

\twocolumn[{
 \maketitle
 \vspace{-2em}

 \includegraphics[width=\textwidth]{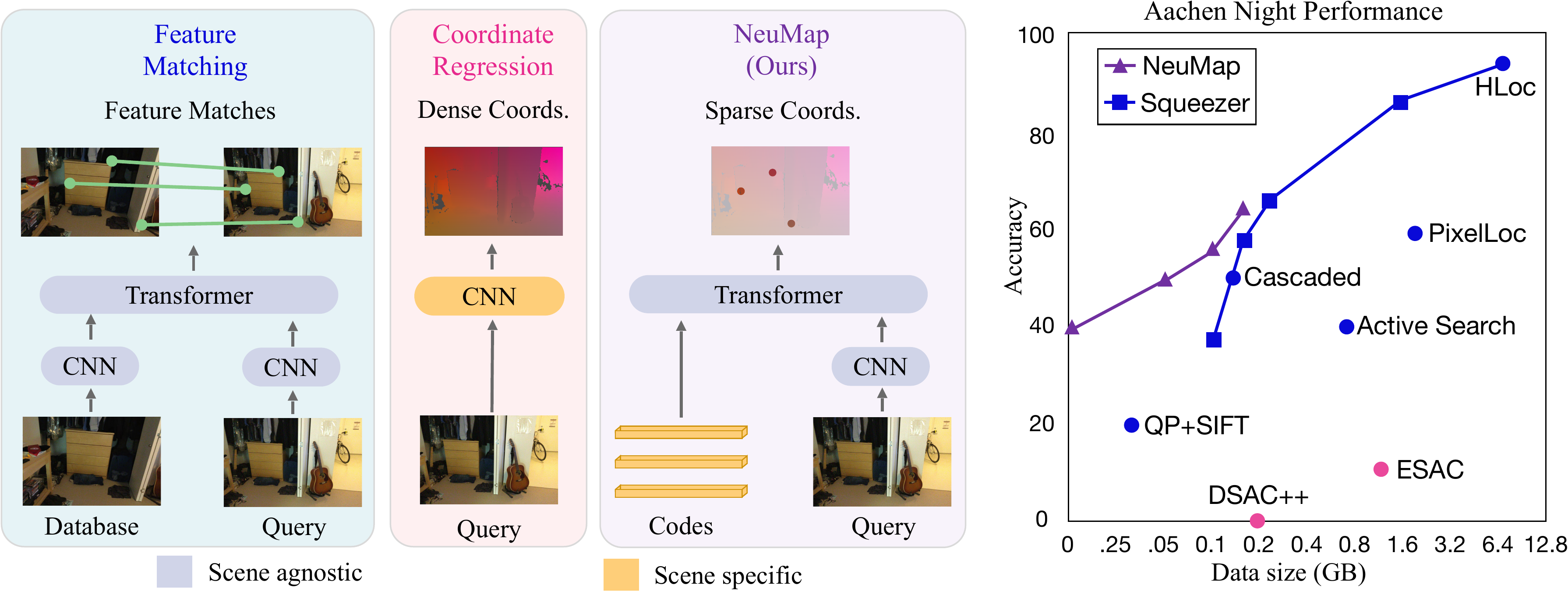} 
\captionof{figure}{
The paper presents neural coordinate mapping (NeuMap) for camera localization. NeuMap encodes a scene into a set of codes and uses a scene-agnostic transformer to decode the coordinates of key-points in a query image. The right compares the localization accuracy and the data size for Aachen Night benchmark.
The accuracy is averaged over three translation/rotation error thresholds (0.25m, 2$^{\circ}$), (0.5m, 5$^{\circ}$), or (5m, 10$^{\circ}$). NeuMap and Squeezer control representation sizes and are illustrated by curves with multiple data points.
%
}
\vspace{1em}
\label{fig:teaser}
}]

\begin{abstract}

This paper presents an end-to-end neural mapping method for camera localization, dubbed NeuMap, encoding a whole scene into a grid of latent codes, with which a Transformer-based auto-decoder regresses 3D coordinates of query pixels. State-of-the-art feature matching methods require each scene to be stored as a 3D point cloud with per-point features, consuming several gigabytes of storage per scene. While compression is possible, performance drops significantly at high compression rates. Conversely, coordinate regression methods achieve high compression by storing scene information in a neural network but suffer from reduced robustness. NeuMap combines the advantages of both approaches by utilizing 1) learnable latent codes for efficient scene representation and 2) a scene-agnostic Transformer-based auto-decoder to infer coordinates for query pixels. This scene-agnostic network design learns robust matching priors from large-scale data and enables rapid optimization of codes for new scenes while keeping the network weights fixed. Extensive evaluations on five benchmarks show that NeuMap significantly outperforms other coordinate regression methods and achieves comparable performance to feature matching methods while requiring a much smaller scene representation size. For example, NeuMap achieves 39.1\% accuracy in the Aachen night benchmark with only 6MB of data, whereas alternative methods require 100MB or several gigabytes and fail completely under high compression settings. The codes are available at \href{https://github.com/Tangshitao/NeuMap}{https://github.com/Tangshitao/NeuMap}.

\end{abstract}

\input{tex_file/1_introduction}
\input{tex_file/2_relatedwork}
\input{tex_file/3_method}

\input{tex_file/4_experiments}

\input{tex_file/5_conclusion}

\newpage
\clearpage

{
    \small
    \bibliographystyle{ieee_fullname}
    \bibliography{egbib}
}

\end{document}

%% file: preamble.tex
\usepackage{xcolor}
\usepackage{xspace}
\usepackage{mathtools}

\definecolor{turquoise}{cmyk}{0.65,0,0.1,0.3}
\definecolor{purple}{rgb}{0.65,0,0.65}
\definecolor{dark_green}{rgb}{0, 0.5, 0}
\definecolor{orange}{rgb}{0.95, 0.6, 0.05}
\definecolor{red}{rgb}{0.8, 0.2, 0.2}
\definecolor{darkred}{rgb}{0.3, 0.05, 0.025}
\definecolor{blueish}{rgb}{0.0, 0.3, .6}
\definecolor{light_gray}{rgb}{0.7, 0.7, .7}
\definecolor{pink}{rgb}{1, 0, 1}
\definecolor{greyblue}{rgb}{0.25, 0.25, 1}

\newcommand{\TODO}[1]{\textbf{\color{red}[TODO: #1]}}

\usepackage{enumitem}
\setlist[itemize]{noitemsep,leftmargin=*,topsep=0em}
\setlist[enumerate]{noitemsep,leftmargin=*,topsep=0em}
\setlist[description]{noitemsep,leftmargin=.1in,topsep=0em}

\renewcommand{\paragraph}[1]{\noindent\textbf{#1}.}

\usepackage{lipsum}

\usepackage{currfile}

\usepackage{environ}
\NewEnviron{scaletoline}{\resizebox{1.0\linewidth}{!}{\BODY}}

\newcommand{\loss}[1]{\mathcal{L}_\text{#1}}

\newcommand{\scene}{\mathcal{S}}
\newcommand{\iScenes}{s}

\newcommand{\hyper}{\lambda}
\newcommand{\W}{\mathbf{W}}
\newcommand{\I}{\mathbf{I}}

\newcommand{\camera}{\mathbf{T}}
\newcommand{\position}{\mathbf{x}}

\newcommand{\keypoint}{\mathbf{k}}
\newcommand{\scenepoint}{\mathbf{K}}

\newcommand{\featureBackbone}{\mathcal{F}}
\newcommand{\feature}{\mathbf{f}}

\newcommand{\voxel}{\mathcal{V}}
\newcommand{\gridEdgeLength}{l}

\newcommand{\tokens}{\mathbf{S}}
\newcommand{\token}{\mathbf{s}}
\newcommand{\iTokens}{t}

\newcommand{\confidence}{c}
\newcommand{\decoder}{\mathcal{D}}

%% file: tex_file/1_introduction.tex
\section{Introduction}

Visual localization determines camera position and orientation based on image observations, an essential task for applications such as VR/AR and self-driving cars. Despite significant progress, accurate visual localization remains a challenge, especially when dealing with large viewpoint and illumination changes. Compact map representation is another growing concern, as applications like delivery robots may require extensive maps. Standard visual localization techniques rely on massive databases of keypoints with 3D coordinates and visual features, posing a significant bottleneck in real-world applications.


Visual localization techniques generally establish 2D-3D correspondences and estimate camera poses using perspective-n-point (PnP)~\cite{lepetit2009epnp} with a sampling method like RANSAC~\cite{fischler1981random}. These methods can be divided into two categories: Feature Matching (FM)~\cite{sarlin2019coarse, sattler2012improving, cheng2019cascaded, yang2022scenesqueezer} and Scene Coordinate Regression (SCR)~\cite{brachmann2017dsac, brachmann2019expert, shotton2013scene, brachmann2021visual}. FM methods, which are trained on a vast amount of data covering various viewpoint and illumination differences, use sparse robust features extracted from the query image and matched with those in candidate scene images. This approach exploits learning-based feature extraction and correspondence matching methods~\cite{detone2018superpoint, sarlin2020superglue, sun2021loftr, tang2022quadtree} to achieve robust localization. However, FM methods require large maps, making them impractical for large-scale scenes. Many methods~\cite{yang2022scenesqueezer, cheng2019cascaded, mera2020efficient} have been proposed to compress this map representation, but often at the cost of degraded performance. On the other hand, SCR methods directly regress a dense scene coordinate map using a compact random forest or neural network, providing accurate results for small-scale indoor scenes. However, their compact models lack generalization capability and are 
often restricted to limited viewpoint and illumination changes.
Approaches such as ESAC~\cite{brachmann2019expert} handle larger scenes by dividing them into smaller sub-regions, but still struggle with large viewpoint and illumination changes.

We design our method to enjoy the benefits of compact scene representation of SCR methods and the robust performance of FM methods. Similar to FM methods, we also focus on a sparse set of robustly learned features to deal with large viewpoint and illumination changes. On the other hand, we exploit similar ideas to SCR methods to regress the 3D scene coordinates of these sparse features in the query images with a compact map representation. Our method, dubbed neural coordinate mapping (NeuMap), first extracts robust features from images and then applies a transformer-based auto-decoder (i.e., auto-transdecoder) to learn: 1) a grid of scene codes encoding the scene information (including 3D scene coordinates and feature information) and 2) the mapping from query image feature points to 3D scene coordinates. At test time, given a query image, after extracting image features of its key-points, the auto-transdecoder regresses their 3D coordinates via cross attention between image features and latent codes.
In our method, the robust feature extractor and the auto-transdecoder are scene-agnostic, where only latent codes are scene specific. This design enables the scene-agnostic parameters to learn matching priors across scenes while maintaining a small data size. To handle large scenes, we divide the scene into smaller sub-regions and process them independently while applying a network pruning technique~\cite{liu2017learning} to drop redundant codes.

We demonstrate the effectiveness of NeuMap with a diverse set of five benchmarks, ranging from indoor to outdoor and small to large-scale: 7scenes (indoor small), ScanNet (indoor small), Cambridge Landmarks (outdoor small), Aachen Day $\&$ Night (outdoor large), and NAVER LABS (indoor large). In small-scale datasets (i.e., 7scenes and Cambridge Landmarks), NeuMap compresses the scene representation by around 100-1000 times without any performance drop compared to DSAC++. In large-scale datasets (i.e., Aachen Day $\&$ Night and NAVER LABS), NeuMap significantly outperforms the current state-of-the-art at high compression settings, namely, HLoc~\cite{sarlin2019coarse} with a scene-compression technique~\cite{yang2022scenesqueezer}. In ScanNet dataset, we demonstrate the quick fine-tuning experiments, where we only optimize the codes for a new scene while fixing the scene-agnostic network weights. 

%% file: tex_file/2_relatedwork.tex
\section{Related work} 

\paragraph{Visual localization with feature matching (FM)}
FM-based localization has achieved state-of-the-art performance~\cite{sarlin2019coarse, sarlin2020superglue, detone2018superpoint, sattler2012improving, sun2021loftr, dusmanu2019d2, yang2022scenesqueezer, revaud2019r2d2}. Torsten et al.\cite{sattler2012improving} match query images exhaustively with all 3D points in structure-from-motion models. However, as scenes grow larger, this matching becomes ambiguous, compromising localization robustness. Paul-Edouard\cite{sarlin2020superglue} proposes a coarse-to-fine strategy: 1) coarsely localizing query images using a global image feature~\cite{arandjelovic2016netvlad}, and 2) computing camera poses through local key-point matches. This pipeline has demonstrated significant improvements, with most follow-up works focusing on enhancing feature matching capability~\cite{yang2022scenesqueezer, sarlin2020superglue, sun2021loftr, zhang2022rendernet}. SuperGlue~\cite{sarlin2020superglue} utilizes Transformers to match key-point sets, achieving impressive results. LoFTR~\cite{sun2021loftr} and its variants~\cite{tang2022quadtree,chen2022aspanformer} propose dense matching frameworks without key-points. Nevertheless, these methods require storing large databases of key-point features or images. SceneSqueezer~\cite{yang2022scenesqueezer} offers a compression mechanism for FM-based methods by removing redundant images and key-points, but performance drops significantly at high compression settings. Our approach employs latent codes to store and retrieve key-point information, naturally achieving high compression rates.

\paragraph{Scene coordinate regression (SCR)}
Convolutional neural networks have been used to regress dense coordinate maps from RGB images for localization~\cite{brachmann2017dsac, brachmann2019neural, brachmann2021visual, brachmann2019expert, li2020hierarchical, wu2022sc}. Due to limited network capacity, this approach does not scale well to large scenes. ESAC~\cite{brachmann2019expert} addresses scalability by dividing large scenes into smaller sub-regions, training a separate network for each. HSCNet~\cite{li2020hierarchical} classifies pixels into corresponding sub-regions to reduce ambiguity. Although SCR-based methods have a smaller representation size than FM-based methods, they are less robust against large viewpoint changes or illumination differences. This is because acquiring SCR training data is expensive, whereas FM training data is easily accessible from extensive databases of feature correspondences in challenging conditions~\cite{dai2017scannet, li2018megadepth}. Furthermore, SCR approaches require model training for each new scene, which is undesirable in many applications. While SANet~\cite{yang2019sanet} and DSM~\cite{tang2021learning} generalize this approach to unseen scenes, they also result in reduced localization accuracy. Our approach is SCR-based, utilizing a scene-agnostic feature extractor and transformer trained as in FM-based methods, offering both robustness and compact representation.

\paragraph{Auto-decoder}
An auto-decoder optimizes latent codes directly via back-propagation within a decoder-only framework~\cite{park2019deepsdf, bojanowski2017optimizing, sitzmann2019scene}. Piotr et al.~\cite{bojanowski2017optimizing} employ an auto-decoder for generative adversarial networks. DeepSDF~\cite{park2019deepsdf} and SRN~\cite{sitzmann2019scene} store shape representations into latent codes. However, MLP or CNN-based auto-decoders have limited capacity using a single code. Transformer-based auto-decoders (i.e., auto-transdecoders) employ an arbitrary number of codes to increase capacity. Recently, Sandler et al.~\cite{sandler2022fine} augment a transformer with learnable memory tokens (codes) that allow the model to adapt to new tasks while optionally preserving its capabilities on previously learned tasks. The transformer weights are first pretrained with a large-scale dataset, and the tokens (codes) are then fine-tuned in downstream tasks. This memory mechanism resembles our auto-transdecoder, but we optimize both transformer and codes simultaneously during training. To the best of our knowledge, this work is the first to use an auto-transdecoder for mapping and localization tasks.

%% file: tex_file/3_method.tex
\input{fig/overview.tex}

\section{Method}

Given a scene represented by a set of reference images~$\{\I_n \}$ with known camera calibrations $\{\camera_n \}$ expressed in a (scene-specific) coordinate frame, we devise a technique that encodes these images into a \textit{compact} scene representation $\scene$. We employ this scene representation to perform \textit{visual localization}, a task of predicting the camera $\camera_q$ of a query image~$\I_q$ that was never seen before -- with a vantage point and/or appearance different from the one in the reference image set.

\subsection{Overview}
We achieve visual localization by solving the proxy task of \textit{scene coordinate regression} on \textit{sparse features}, where given a set of 2D key-point~$\{ \keypoint_i \}$ extracted from $\I_q$, we predict its corresponding 3D scene coordinate $\{ \scenepoint_i \}$~(\autoref{sec:regression}).
As shown in Figure~\ref{fig:overview}, our method extracts 2D key-points $\{ \keypoint_i \}$ following HLoc~\cite{sarlin2019coarse} with a pre-trained backbone, R2D2~\cite{revaud2019r2d2}. 
In order to determine their scene coordinates, we first compute a feature map of the image $\I_q$ with a trainable CNN backbone $\featureBackbone$, which is a ResNet18~\cite{he2016deep} and bilinearly interpolates the feature, and then solve the scene coordinate by a decoder $\decoder$ as,

\begin{equation}
\scenepoint_i = \decoder(\featureBackbone(\I_q, \keypoint_i), \mathcal{S})
\label{eq:sfr}
\end{equation}
Here, $\mathcal{S}$ is the learned scene representation. Finally, we obtain camera localization by solving a perspective-n-point (PnP) problem~\cite{lepetit2009epnp}~(\autoref{sec:inference}) with 2D-3D correspondences from $\{ \keypoint_i \leftrightarrow \scenepoint_i \}$.

\subsection{Sparse Scene Coordinate Regression}
\label{sec:regression}

\paragraph{Scene Representation}
We first build a scene model to facilitate localization. Given a set of reference images, we extract 2D key-points by R2D2~\cite{revaud2019r2d2} and compute their corresponding 3D scene coordinates by COLMAP~(\autoref{sec:training}). Some of the key-points are not successfully triangulated, so their 3D coordinates are invalid. We cache 3D scene points in a \textit{sparse grid} with voxels of uniform side length~$\gridEdgeLength$ as shown in Figure~\ref{fig:overview}. 
We denote the $v$-th voxel as $\voxel_v$.
We then model the reference scene as a \textit{set} of latent codes $\scene {=} \{ \tokens_v \}$ -- \textit{one per voxel} -- and modify Equation~\ref{eq:sfr} to reflect this design choice as:
\begin{equation}
(\scenepoint^v_i, c^v_i) = \decoder(\featureBackbone(\I_q, \keypoint_i), \tokens_v)
\label{eq:hybrid-notsparse}
\end{equation}
Here, the scalar parameter $c^v_i$ is the confidence that the key-point $\keypoint_i$ is in the voxel $\voxel_v$, and $\scenepoint^v_i$ is the scene coordinate under the voxel attached coordinate frame, i.e., $\scenepoint_i = \scenepoint^v_i + \ve{O}_v$ and $\ve{O}_v$ is the mean of all the coordinates in $\voxel_v$. 

This formulation simultaneously solves a classification and regression problem: 1) classifying whether a 2D key-point belongs to a 3D voxel and 2) regressing its 3D position \textit{locally} within the voxel.

\paragraph{Sparsity}
Inspired by network pruning~\cite{liu2017learning}, we apply code pruning to remove redundant codes. Specifically, we multiply each voxel code in $\tokens_v$ with a scaling factor $w_v$, and jointly learn the codes with these scaling factors, with $L1$ loss imposed on $w_v$. After finishing training, we prune codes whose weights are below a threshold and finetune the remaining codes.

\paragraph{Scene Coordinate Regression}
\label{sec:decoder}
We use a set of per-voxel latent codes $\tokens_v$ to facilitate the learning of scene coordinate regression. 
The decoder $\mathcal{D}$ is a stacked transformer to regress the scene coordinates of the 2D image key-points. We include $T$ transformer blocks, (6 blocks in our implementation), defined by the inductive relationship~(see Figure ~\ref{fig:overview}) as,
\begin{align}
\feature_i^{(t+1)} &= \text{CrAtt}(\feature_i^t, \Tilde{\token}_v^t) \\
\Tilde{\token}_v^t &= w_v^t \token_v^t
\label{eq:attention}
\end{align}
where the feature $\feature_i^{(1)}{=}\featureBackbone(\I_q, \keypoint_i)$, $w_v^t$ is the scaling factor enforcing sparsity. Each transformer block contains a set of codes $\token_v^t$, and the final per voxel codes are~$\tokens_v {=} \{\token_v^t, 1 {\leq} t {\leq} T \}$.  The function $\text{CrAtt}(\cdot, \cdot)$ is classical cross-attention from transformers~\cite{attention}:
\begin{gather}
\mathbf{Q} = \W_\mathbf{Q}^{(\iTokens)} \cdot \feature^t_i, 
\mathbf{K} = \W_\mathbf{K}^{(\iTokens)} \cdot \Tilde{\token}_v^t, 
\mathbf{V} = \W_\mathbf{V}^{(\iTokens)} \cdot \Tilde{\token}_v^t
\label{eq:QKV}
\\
\feature^{(t+1)}_i = \text{MLP}(\text{softmax}(\mathbf{Q} \cdot \mathbf{K}) \cdot \mathbf{V})
\end{gather}
At the end of the stacked transformer, we apply another MLP to compute the scene coordinate and confidence as,
\begin{equation}
(\scenepoint^v_i, c^v_i) = \text{MLP}(\feature_i^{(T)})
\end{equation}

\subsection{Training}
\label{sec:training}

At training time, we learn the weights of the feature encoder $\featureBackbone$, the decoder $\decoder$, and the \textit{compressed} scene representation for all training scenes~$\{\scene_\iScenes\}$ via generative latent optimization / auto-decoding ~\cite{park2019deepsdf} with the following loss:
\begin{align}
\loss{} = \hyper_\position \cdot \loss{$\position$} + \hyper_\confidence \cdot \loss{$\confidence$} +  \hyper_\text{L1} \cdot \loss{L1}.
\label{eq:training}
\end{align}
We set the loss weights $\hyper_\text{x}$, $\hyper_\text{c}$, and $\hyper_\text{L1}$ to 1.

The scene coordinate loss is defined as,
\begin{equation}
\loss{$\position$} = 
\sum_v \sum_i \:\:
 (y_i^v) \cdot \| (\scenepoint^v_i + \ve{O}_v) - \scenepoint^*_i \|_2
\end{equation}
where $\scenepoint^*_i$ is the scene coordinate triangulated by COLMAP for key-point $\keypoint_i$, the binary variable $y_i^v$ indicates whether the triangulated scene coordinate $\scenepoint^*_i$ belongs to the voxel~$\voxel_v$. The key-points with no valid 3D coordinates (not successfully triangulated) do not belong to any voxel.

The second term is a classification loss, i.e., a binary cross entropy, for the confidence $c_i^v$,
\begin{align}
\loss{$\confidence$} =  \sum_v\sum_i BCE(c^v_i, y^v_i) \end{align}

The third term enforces \textit{sparsity} and produces a \textit{compressed} representation, which is defined as,
\begin{align}
\loss{L1} &=  \sum_{v} \sum_{t=1}^T \| w_v^t \|_1,
\end{align}
where $w_v^t$ is the sparsity factor in Equation~\ref{eq:attention}.

\paragraph{Training strategy}
We learn the decoder $\decoder$, the CNN backbone $\featureBackbone$, and the scene representation $\scene$ with voxel sampling. At each iteration, we randomly choose $B$ voxels, where $B$ is the batch size. Each voxel $\voxel_v$ has a set of reference images $\ve{I}_v$, each of which contains at least 20 scene points in $\voxel_v$. We then sample one training image for each voxel and optimize $\decoder$, $\featureBackbone$, and the scene codes of sampled voxels by minimizing the training loss in Equation~\ref{eq:training}. We sample voxels without replacement, so all scene codes are updated once at each epoch.

Similarly to network pruning~\cite{liu2017learning}, we minimize the training loss to convergence, set sparsity factors $w_v^t$ whose values are below a certain threshold to zero, and fine-tune our model while keeping $w_v^t$ frozen.

\paragraph{Scene adaption}
Given a \textit{new} scene (i.e., not in the training data), we simply optimize the scene code $\scene$, while keeping decoder $\decoder$ and CNN backbone $\featureBackbone$ \textit{fixed}. In this way, our scene representation $\scene$ is scene specific, but the decoder $\decoder$ and feature extractor $\featureBackbone$ are scene agnostic.

\subsection{Inference}
\label{sec:inference}
Given a query image $\I_q$, we use an existing deep image retrieval method~\cite{revaud2019learning, arandjelovic2016netvlad} to retrieve the most similar reference images, which activate a subset of voxels; see Figure~\ref{fig:overview}. A voxel $\voxel_v$ is activated if one of the retrieved reference images contains at least 20 scene points in $\voxel_v$. For large-scale scenes, we typically get around 100-200 voxels, while for small-scale scenes, we consider all the voxels without image retrieval. 
We then extract a set of 2D key-points $\{\keypoint_i \}$ within $\I_q$, and for each of them, regress their per-voxel confidence and positions via Equation~\ref{eq:hybrid-notsparse}. We discard points with confidence $\confidence{<}0.5$. All the remaining points are used to compute the camera pose with the PnP algorithm combined with RANSAC, implemented in Pycolmap~\cite{pycolmap2022}.

%% file: fig/overview.tex
\begin{figure*}[t!]
\centering

\includegraphics[width=\linewidth]{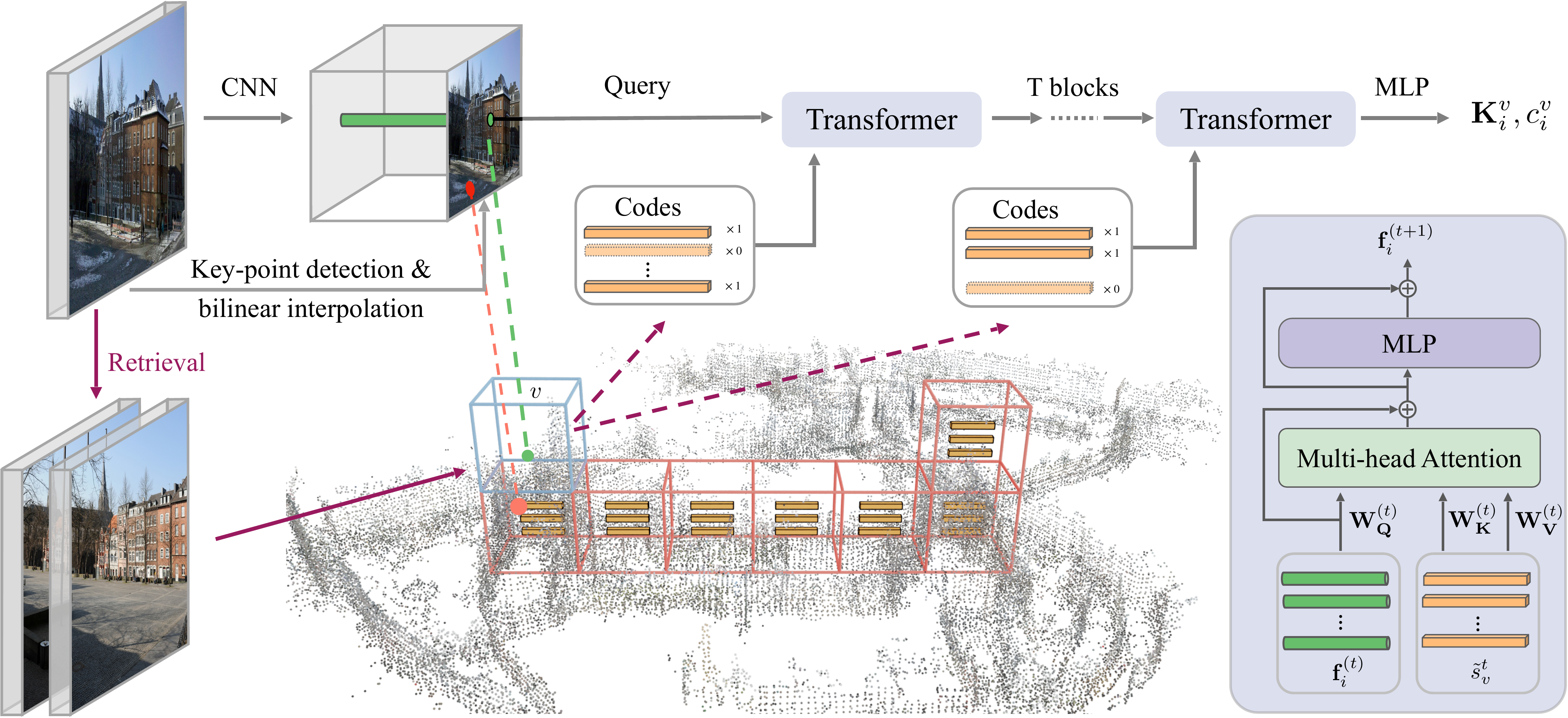}
\caption{\textbf{Overview} -- We divide a scene into sub-regions and assign codes to each region. A shared convolutional neural network extracts image features, and a Transformer network decodes coordinates from the codes. Code-pruning further reduces the data size.
}
\label{fig:overview}
\end{figure*}

%% file: tex_file/4_experiments.tex
\section{Experiments} \label{section:experiments}
We conduct extensive evaluations with twelve competing methods on five benchmarks.

\begin{figure*}
    \centering
    \small
    \includegraphics[width=0.95\textwidth]{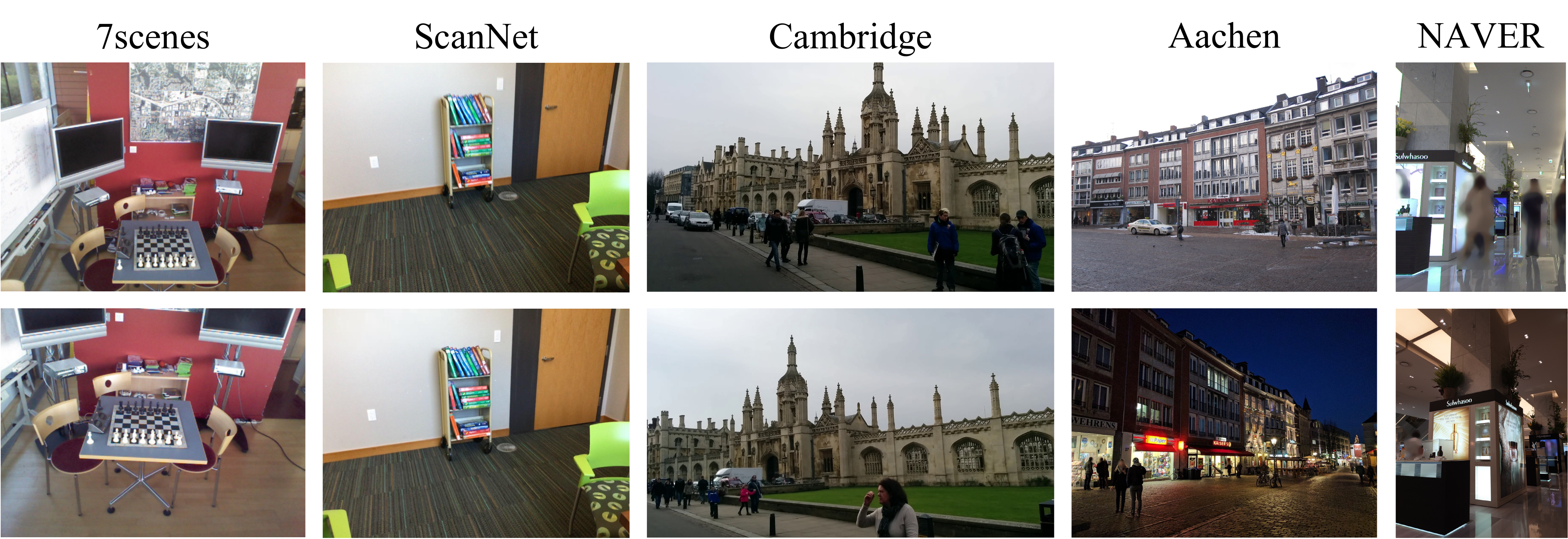}
    \caption{Sample images from 7scenes, ScanNet, Cambridge, Aachen Day $\&$ Night, and NAVER LABS datasets.}
    \label{fig:dataset}
\end{figure*}

\subsection{Datasets}
We use 7scenes~\cite{shotton2013scene}, Cambridge landmarks~\cite{kendall2015posenet}, Aachen Day $\&$ Night~\cite{sattler2012image}, NAVER LABS datasets~\cite{lee2021large}, and ScanNet~\cite{dai2017scannet} (See Fig.~\ref{fig:dataset}). For 7scenes, Cambridge, Aachen, and ScanNet, we train a separate model for each dataset. Within each dataset, all the scenes share the same network parameters. For NAVER, we train one model for one scene due to the large size of the scene codes, which cannot fit in GPU memories.

Additionally, for 7scenes, Cambridge, and NAVER, we train models from scratch. For Aachen, with images that have significant view and illumination differences, we use a feature extractor from LoFTR~\cite{sun2021loftr}, pretrained with images containing such variations. Although ScanNet is not a standard localization dataset, it shares similarities with 7scenes. To illustrate our scene-agnostic network and scene-specific code design, we use 80 scenes to train the scene-agnostic network and fine-tune the codes on 1 scene (scene0708) for ScanNet.

We demonstrate code pruning for 7scenes, Cambridge, and Aachen, with pruning thresholds of 0.00003, 0.05, and 0.2, respectively. We use median translation/rotation errors to evaluate 7scenes, Cambridge, and ScanNet. For Aachen and NAVER, the accuracy under different thresholds is the standard metric used.

\subsection{Implementation details} We employ ResNet-18~\cite{he2016deep} to extract image features and train models with 8 V100 GPUs using a batch size of 256. We set the initial learning rate to 0.002 for training the scene-agnostic parameters and 0.0001 for training the scene-specific codes. After every 30 epochs, we multiply the learning rate by 0.5. We train the model for 200 epochs in the first stage and fine-tune the model for 100 epochs in the second stage. For training data generation, we use r2d2~\cite{revaud2019r2d2} to extract keypoints and perform triangulation to obtain coordinates in Cambridge, Aachen, and NAVER. We use depth images to acquire 3D coordinates in 7scenes and ScanNet.

\subsection{Main results with code-pruning}
We present the results after code-pruning for 7scenes, Cambridge, and Aachen datasets.

\input{table/Aachen_table}
\input{table/7scene_table}
\input{table/cambridge_table}
\paragraph{Aachen Day $\&$ Night}
Table ~\ref{tab:aachen} compares NeuMap with other methods on this large-scale outdoor dataset. We implement three versions of NeuMap by varying the size of voxels (either 8 or 10 meters) with or without code-pruning. The version with 8 meters without code-pruning serves as the performance upper-bound.

Existing coordinate regression methods (i.e., ESAC or DSAC++) perform poorly on this dataset. DSAC++ cannot fit the whole scene into a single network. ESAC is not robust to view and illumination differences. NeuMap is the first coordinate regression method with competitive results in large-scale outdoor datasets. NeuMap also outperforms another non-feature matching-based method, PixLoc, by a significant margin. Cascaded~\cite{cheng2019cascaded}, QP+R.Sift~\cite{mera2020efficient}, and Squeezer~\cite{yang2022scenesqueezer} are scene compression methods based on feature matching. NeuMap outperforms Cascaded and QP+R.Sift by a considerable margin. Both methods utilize numeric quantization for compression (converting 32-bit floats to 8-bit integers), which is orthogonal to our scheme and would likely further improve our performance once combined.

Squeezer~\cite{yang2022scenesqueezer} is built on HLoc~\cite{sarlin2019coarse}, which extracts and matches keypoints using SuperPoint~\cite{detone2018superpoint} and SuperGlue~\cite{sarlin2020superglue}. Squeezer removes redundant keypoints by solving quadratic programming and achieves impressive compression at the expense of performance. We use SuperPoint~\cite{detone2018superpoint} features (float32) for Squeezer. The NeuMap with code pruning (10 meters) achieves similar performance as Squeezer with a 70 Megabytes smaller representation size. Even at extremely high compression ratios, NeuMap obtains good results, where the total mapping size is only 6 Megabytes. Squeezer fails at such ratios, as shown in Fig.~\ref{fig:teaser}.

\paragraph{Cambridge} 
Table ~\ref{tab:cambridge} compares performance on the Cambridge dataset. We use triangulated points for supervision and divide each scene into voxels of 200 meters in length. NeuMap achieves similar performance as DSAC++ with a 100 to 1000 times smaller representation size. Our compression design is simple (i.e., latent codes with auto-transdecoder) compared to Squeezer, which has a complex pipeline.

\paragraph{7scenes} 
Table ~\ref{tab:7scenes} shows the results, where we use depth images for supervision and train models from scratch. A scene is divided into voxels of 3 meters in length. NeuMap achieves the same performance as DSAC++, HLoc, and DSM with a 200-1000 times smaller data size.

\subsection{Main results without code-pruning}

\input{table/nl_table}
NAVER is a large-scale indoor localization dataset. Each scene contains approximately 20,000 images. We use 3 scenes (Dept. B1), (Dept. 1F), and (Dept. 4F) for experiments. We divide each scene into voxels of 4 meters in length, which results in approximately 2,000 voxels, and assign $50\times 6$ codes to each voxel, and train a model from scratch without code pruning. Table ~\ref{table:naver} compares NeuMap against three methods. Each method is trained with "training images" and evaluated with "validation images" of their dataset. NeuMap outperforms the state-of-the-art coordinate regression method ESAC by a large margin with a much smaller representation size. In comparison to feature matching methods (D2Net and R2D2), NeuMap has similar accuracy when the error thresholds are (0.25m, 2$^{\circ}$) and (1m, 5$^{\circ}$), while the representation size is 200 times more compact even without code pruning.

\subsection{Evaluating code fine-tuning}

\input{table/scannet_table}
\begin{figure*}
    \centering
    \includegraphics[width=\textwidth]{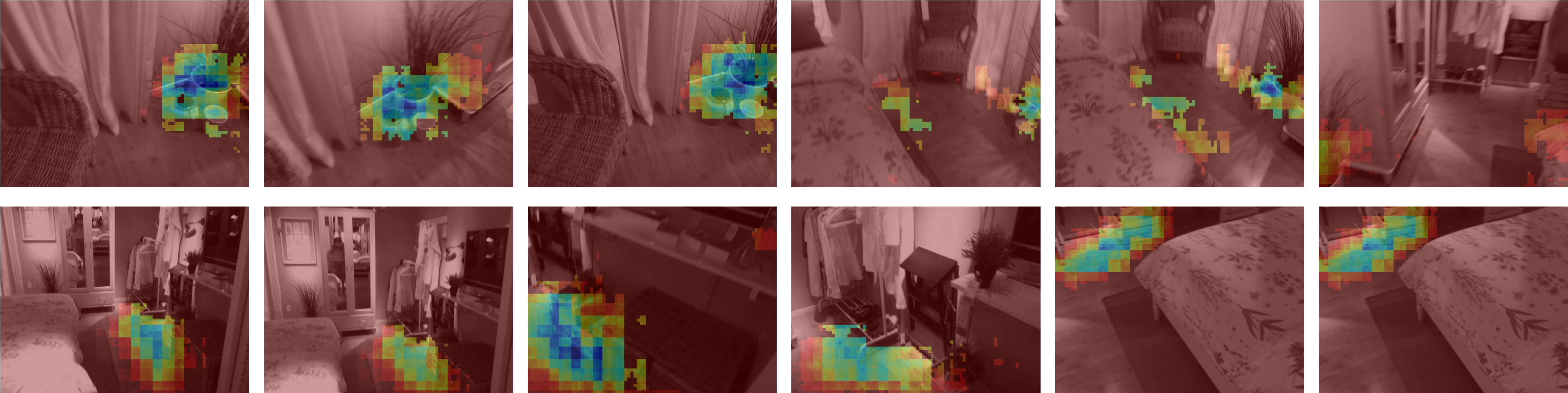}
    \caption{Attention score visualization. We choose one specific code, compute the attention scores for all pixels, and visualize scores by normalizing all the pixels. Each row is the visualization results for the same code. We can see that each code memorizes a specific area. } \label{fig:attn_score}
\end{figure*}
Our scene-agnostic network and scene-specific code design allows us to fine-tune only the codes for a new scene while fixing the network weights. We use the ScanNet dataset for experiments. We train NeuMap with 1, 10, 20, 40, or 80 scenes and choose 1 new testing scene for fine-tuning and evaluation. For each scene, we randomly sample 10\% of the frames for testing, while the rest becomes the training set.

Figure ~\ref{fig:scannet} shows the median translation and the median rotation errors for the fully-trained (i.e., optimizing both network weights and codes) and the fine-tuned NeuMap models. The fully-trained models' accuracy does not drop as the number of training scenes increases, showing that NeuMap handles an arbitrary number of scenes without a performance drop. On the other hand, fine-tuned models improve accuracy as more training scenes are used to learn a generalizable scene-agnostic network.

\subsection{Ablation studies and visualization}

We conduct several ablation studies and show visualization results for NeuMap.

\input{table/code_table}
\paragraph{Code number and code-pruning} 
Table ~\ref{tab:code} shows how the number of codes influences the accuracy of the Aachen Day $\&$ Night dataset. NeuMap obtains good results even with 1 code, while the performance improves as we add more codes. The table also shows the results of code pruning, which reduces the number of codes from 100 to 25 on average while maintaining similar accuracy. The pruned version with 25 codes is far superior to the counterpart without pruning, which was trained with 25 codes from the start.

\input{table/block_table}

\paragraph{Attention score visualization} Figure ~\ref{fig:attn_score} shows the attention scores between one code and image features in different views. The score for pixel $i$ and code $j$ is computed as
\begin{align}
    s_i^j = [\text{softmax}(\mathbf{Q} \cdot \mathbf{K})]^j.
\end{align}
$[.]^j$ is the $j^{th}$ element of the vector. $\mathbf{Q}$, $\mathbf{K}$ is the same as Equation \ref{eq:QKV}. We normalize the scores across an image by
\begin{align}
    \Tilde{s}_i^j=\frac{((s_i^j)-a)}{a-b}.
\end{align}
$a$ and $b$ are the minimum and maximum values in an image. A code activates and stores information for consistent pixels corresponding to the same 3D scene structures.

\begin{figure}
    \centering
    \includegraphics[width=0.48\textwidth]{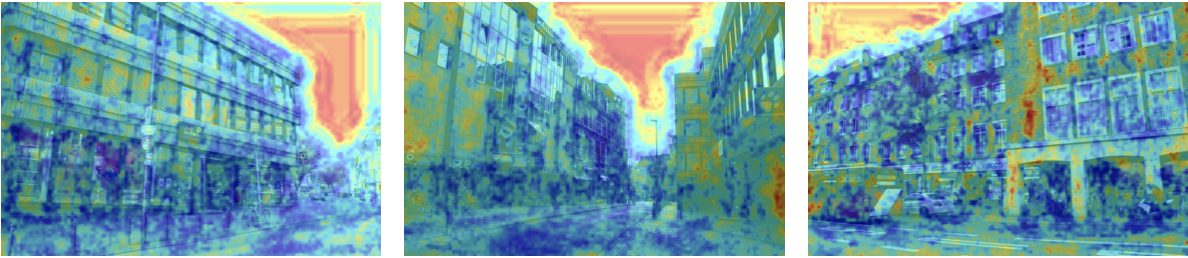}
    \caption{Visualization of activated regions. Deep blue delineates larger attention scores, while orange corresponds to smaller scores.}
    \label{fig:correlated_region}
\end{figure}
\paragraph{Correlated region visualization}
In Fig.~\ref{fig:correlated_region}, we visualize the most correlated regions using a heat map. We computed attention scores for each feature point on a dense feature map. The visualization score for feature point $i$ is determined by the maximum score among all codes, expressed as $\max_{j} s_i^j$, where $j$ represents the codes. After normalizing the scores across the entire feature map, we blended them with the RGB image for a comprehensive visualization. This heat map highlights the most significant regions in the scene.

\paragraph{Transformer block number} Table ~\ref{tab:block_num} shows the results under different transformer block numbers. The performance improves as we use more transformer blocks.

%% file: table/Aachen_table.tex
\begin{table}[tb]
\small
\centering
\scalebox{0.95}{\begin{tabular}{@{\hskip1pt}c@{\hskip5pt}c|c|c|c}
              && Size  & Aachen Day                                 & Aachen Night                               \\ \hline
\multirow{5}{*}{\begin{turn}{90}FM\end{turn}}  & AS~\cite{sattler2016efficient}            & 0.75  & 57.3 / 83.7 / \textcolor{orange}{96.6}                         & 28.6 / 37.8 / 51.0                         \\
&HLoc~\cite{sarlin2019coarse}          & 7.82 & \textcolor{cyan}{89.6} / \textcolor{cyan}{95.4} / \textcolor{cyan}{98.8}                         & \textcolor{cyan}{86.7} / \textcolor{cyan}{93.9} / \textcolor{cyan}{100}                        \\
&\textcolor{red}{Cascaded}~\cite{cheng2019cascaded}      & 0.14  & 76.7 / 88.6 / 95.8                         & 33.7 / 48.0 / 62.2                         \\
&\textcolor{red}{QP+R.Sift}~\cite{mera2020efficient}     & \textcolor{orange}{0.03}   & 62.6 / 76.3 / 84.7                         & 16.3 / 18.4 / 24.5                         \\
&\textcolor{red}{Squeezer}~\cite{yang2022scenesqueezer} & 0.24   & 75.5 / 89.7 / 96.2                         & 50.0 / 67.3 / 78.6                         \\ \hline
\multirow{5}{*}{\begin{turn}{90}E2E\end{turn}}  & PixLoc~\cite{sarlin2021back}   & 2.13 & 64.3 / 69.3 / 77.4                         & \textcolor{orange}{51.1} / 55.1 / 67.3                         \\
&ESAC(50)~\cite{brachmann2019expert}      & 1.31 & 42.6 / 59.6 / 75.5                         & \hphantom{2}6.1 / 10.2 / 18.4                          \\
&Ours (8m, 100) & 1.26 & \textcolor{orange}{80.8} / \textcolor{orange}{90.9} / 95.6 & 48.0 / \textcolor{orange}{67.3} / \textcolor{orange}{87.8} \\
&Ours (10m, 1) & \textcolor{cyan}{.006} & 49.3 / 75.2 / 93.9 & 14.3 / 26.5 / 76.5 \\
&Ours (10m, *)        & 0.17  & 76.2 / 88.5 / 95.5 & 37.8 / 62.2 / 87.8                    
\end{tabular}}
\caption{Results of Aachen Day $\&$ Night. We report accuracy under different error thresholds. The first number in the bracket is the sub-region length. The second number is the number of codes for each transformer block, and the notation (*) indicates the application of code pruning. Data sizes in the "Size" column are reported in Gigabytes. The \textcolor{red}{red} texts show methods with scene compression. FM means feature matching. E2E means end-to-end. The same applies to the following figures. We outperform all the other E2E methods by a large margin, achieving similar performance as Squeezer, which applies scene compression to HLoc. The \textcolor{cyan}{cyan} and
 \textcolor{orange}{orange} colors indicate the best and the second best methods, respectively, in each column.
}
\label{tab:aachen}
\end{table}

%% file: table/7scene_table.tex
\begin{table*}[tb]
\small
\centering
\begin{tabular}{cc|c|c|c|c|c|c|c|c}

         &7scenes &size (GB) &Chess&Fire&Heads&Office&Pumpkin&Kitchen&Stairs\\
         
         \hline
\multirow{2}{*}{\begin{turn}{90}FM\end{turn}} &Active Search~\cite{sattler2012improving}& $>$0.5&1.96, 0.04&1.53, \textcolor{orange}{0.03}&1.45, \textcolor{orange}{0.02}& 3.61, \textcolor{orange}{0.09}&3.10, 0.08&3.37, 0.07&2.22, \textcolor{cyan}{0.03}\\
         
         &HLoc~\cite{sarlin2019coarse} &$>$1.0 &0.79, \textcolor{cyan}{0.02}&\textcolor{orange}{0.87}, \textcolor{cyan}{0.02}&\textcolor{orange}{0.92}, \textcolor{orange}{0.02}&0.91, \textcolor{cyan}{0.03}& \textcolor{orange}{1.12},  \textcolor{orange}{0.05}&\textcolor{orange}{1.25}, \textcolor{orange}{0.04}&1.62, 0.06\\ \hline
\multirow{4}{*}{\begin{turn}{90}E2E\end{turn}} &DSAC++~\cite{brachmann2018learning}& $>$0.2&\textcolor{cyan}{\textbf{0.5}\hphantom{2}}, \textcolor{cyan}{0.02}&0.9\hphantom{2}, \textcolor{cyan}{0.02}&\textcolor{cyan}{0.8\hphantom{2}}, \textcolor{cyan}{0.01}&\textcolor{cyan}{0.7}\hphantom{2}, \textcolor{cyan}{0.03}&\textcolor{cyan}{1.1\hphantom{2}}, \textcolor{cyan}{0.04}&\textcolor{cyan}{1.1\hphantom{2}}, \textcolor{orange}{0.04}&2.6\hphantom{2}, 0.09\\
        &SANet~\cite{yang2019sanet} & $>$1.0&0.88,\textcolor{orange}{0.03}&1.12, \textcolor{orange}{0.03}&1.48, \textcolor{orange}{0.02}&1.00, \textcolor{cyan}{0.03}&1.21, \textcolor{cyan}{0.04}&1.40, \textcolor{orange}{0.04}&4.59, 0.16 \\
         &DSM~\cite{tang2021learning}&$>$1.0&\textcolor{orange}{0.68}, \textcolor{cyan}{0.02}& \textcolor{cyan}{0.80}, \textcolor{cyan}{0.02}&\textcolor{cyan}{0.80}, \textcolor{cyan}{0.01}&\textcolor{orange}{0.78}, \textcolor{cyan}{0.03}&\textcolor{cyan}{1.11}, \textcolor{cyan}{0.04}&\textcolor{cyan}{1.11}, \textcolor{cyan}{0.03}&\textcolor{orange}{1.16}, \textcolor{orange}{0.04}\\
         
         &Ours&$<$\textcolor{cyan}{0.002}&0.81, \textcolor{cyan}{0.02}&1.11, \textcolor{orange}{0.03}&1.17, \textcolor{orange}{0.02}&0.98, \textcolor{cyan}{0.03}&\textcolor{cyan}{1.11}, \textcolor{cyan}{0.04}&1.33, \textcolor{orange}{0.04}&\textcolor{cyan}{1.12}, \textcolor{orange}{0.04}\\
\end{tabular}
\caption{Results of 7scenes. We report the median translation and rotation errors in ($^{\circ}$, m). NeuMap achieves similar performance as other methods with significantly smaller representation sizes.}\label{tab:7scenes}
\end{table*}

%% file: table/cambridge_table.tex
\begin{table*}[tb]
\centering
\small
\begin{tabular}{@{\hskip1pt}c@{\hskip5pt}c|c|c|c|c|c}
 && ShopFacade          & OldHospital         & College      & Church & Court \\ \hline
\multirow{5}{*}{\begin{turn}{90}FM\end{turn}}  &Active search~\cite{sattler2016efficient}     & (1.12, 0.12) / 38.7 & (1.12, 0.52) / 140\hphantom{2} &  (0.70, 0.57) / 275\hphantom{2}   &  (0.62, 0.22) / 359\hphantom{2} & (0.60, 1.20) / -\hphantom{222} \\ 
&HLoc~\cite{sarlin2019coarse}          &  (\textcolor{cyan}{0.20}, \textcolor{cyan}{0.04}) / 316 & (\textcolor{cyan}{0.30}, \textcolor{cyan}{0.15}) / 1335 & (\textcolor{cyan}{0.20}, \textcolor{cyan}{0.12}) / 1877 &  (\textcolor{cyan}{0.21}, \textcolor{cyan}{0.07}) / 2007 & (\textcolor{orange}{0.16}, \textcolor{orange}{0.11}) / 2295\\ 
& \textcolor{red}{Hybrid}~\cite{camposeco2019hybrid}        & (0.54, 0.19) / \textcolor{orange}{0.16}  &  (1.01, 0.75) / \textcolor{orange}{0.62} & (0.59, 0.81) / \textcolor{orange}{1.01}  &  (0.49, 0.50) / \textcolor{orange}{1.34}  \\
&\textcolor{red}{QP+RootSIFT}~\cite{mera2020efficient}   &  (1.40, 0.72) / 0.41  &  (2.17, 0.90) / 1.1\hphantom{2}  & (1.09, 1.53) / 2.20   & (0.89, 0.56) / 3.30  \\ 
&\textcolor{red}{Squeezer}~\cite{yang2022scenesqueezer} &  (0.38, 0.11) / 1.04 & (0.57, 0.37) / 4.03 & (0.38, 0.27) / 2.4\hphantom{2}  & (0.37, 0.15) / 7.97 \\ \hline
\multirow{4}{*}{\begin{turn}{90}E2E\end{turn}}  &DSAC++~\cite{brachmann2018learning}& (0.3\hphantom{0}, 0.06) / 207   & (0.3\hphantom{0}, 0.2\hphantom{0}) / 207 & (0.3\hphantom{0}, 0.18) / 207 & (0.4\hphantom{0}, 0.13) / 207 & (0.4\hphantom{0}, 0.2\hphantom{0}) / 207 \\ 
&HSCNet~\cite{brachmann2018learning}        & (0.3\hphantom{0}, \textcolor{orange}{0.06}) / 207   & (\textcolor{orange}{0.3\hphantom{0}}, \textcolor{orange}{0.19}) / 207 & (0.3\hphantom{0}, 0.18) / 207 & (\textcolor{orange}{0.30}, \textcolor{orange}{0.09}) / 207& (0.2\hphantom{0}, 0.28) / 207 \\ 
&DSM~\cite{tang2021learning}        & (0.30, \textcolor{orange}{0.06}) /\hphantom{2}27   & (0.23, 0.38) / 105 & (0.35, 0.19) / 143 & (0.34, 0.11) / 174 & (0.43, 0.19) / 218 \\ 
&Ours          & (\textcolor{orange}{0.25}, \textcolor{orange}{0.06}) / \textcolor{cyan}{0.3} & (\textcolor{orange}{0.36}, \textcolor{orange}{0.19}) / \textcolor{cyan}{0.2} & (\textcolor{orange}{0.19}, \textcolor{orange}{0.14}) / \textcolor{cyan}{0.3}  & (0.53, 0.17) / \textcolor{cyan}{0.4} &(\textcolor{cyan}{0.10}, \textcolor{cyan}{0.06}) / \textcolor{cyan}{1.6}
\end{tabular}
\caption{Results of Cambridge. Each cell reports the median translation error, the median rotation error, and the scene representation size in \{($^{\circ}$, m) / MB\}. \textcolor{red}{Red} indicates scene compression methods. NeuMap achieves similar accuracy as coordinate regression and feature matching-based methods but with significantly smaller representation sizes. }\label{tab:cambridge}
\end{table*}

%% file: table/nl_table.tex
\begin{table*}[tb]
\centering

\begin{tabular}{@{\hskip1pt}cc|c|c|c}
                     &           & Dept. B1                   & Dept. 1F                  & Dept. 4F                   \\ \hline
\multirow{2}{*}{\begin{turn}{90}FM\end{turn}}  & D2Net~\cite{dusmanu2019d2}     & (\textcolor{orange}{70.2} / \textcolor{cyan}{78.0} / (\textcolor{orange}{86.1}) / 505GB    & (\textcolor{orange}{83.2} / \textcolor{orange}{89.2} / (\textcolor{orange}{94.5}) / 398GB   & (\textcolor{orange}{72.1} / \textcolor{orange}{85.3} / \textcolor{orange}{98.5}) / 183GB    \\
                     & R2D2~\cite{revaud2019r2d2}      & (\textcolor{cyan}{71.9} / (\textcolor{orange}{77.8} / \textcolor{cyan}{87.9}) / 210GB     & (\textcolor{cyan}{85.8} / (\textcolor{cyan}{89.9} / 94.4) / 166GB   & (\textcolor{cyan}{72.6} / 84.6 / 98.3) / \hphantom{2}76GB     \\ \hline
\multirow{2}{*}{\begin{turn}{90}E2E\end{turn}} & ESAC (50)~\cite{brachmann2019expert} & (\hphantom{2}5.4 /  \hphantom{2}9.1 / 14.2) / \textcolor{orange}{1.3GB}   & (49.7 / 71.5 / 84.1) / \textcolor{orange}{1.3GB}    & (45.2 / 69.9 / 85.1) / 1.3GB \\
                     & Ours      & (46.0 /66.5 / 79.8) / \textcolor{cyan}{0.8GB} & (75.5 / 88.2 / \textcolor{cyan}{95.8}) / \textcolor{cyan}{0.7GB} &   (70.4 /  \textcolor{cyan}{85.4} / \textcolor{cyan}{99.0}) / \textcolor{cyan}{0.4GB}                       
\end{tabular}
\caption{Results of NAVER LAB. We report accuracy under different error thresholds (0.1m, 1$^{\circ}$), (0.25m, 2$^{\circ}$), (1m, 5$^{\circ}$). 
}
\label{table:naver}
\end{table*}

%% file: table/scannet_table.tex

\begin{figure}
    \centering
    \includegraphics[width=0.5\textwidth]{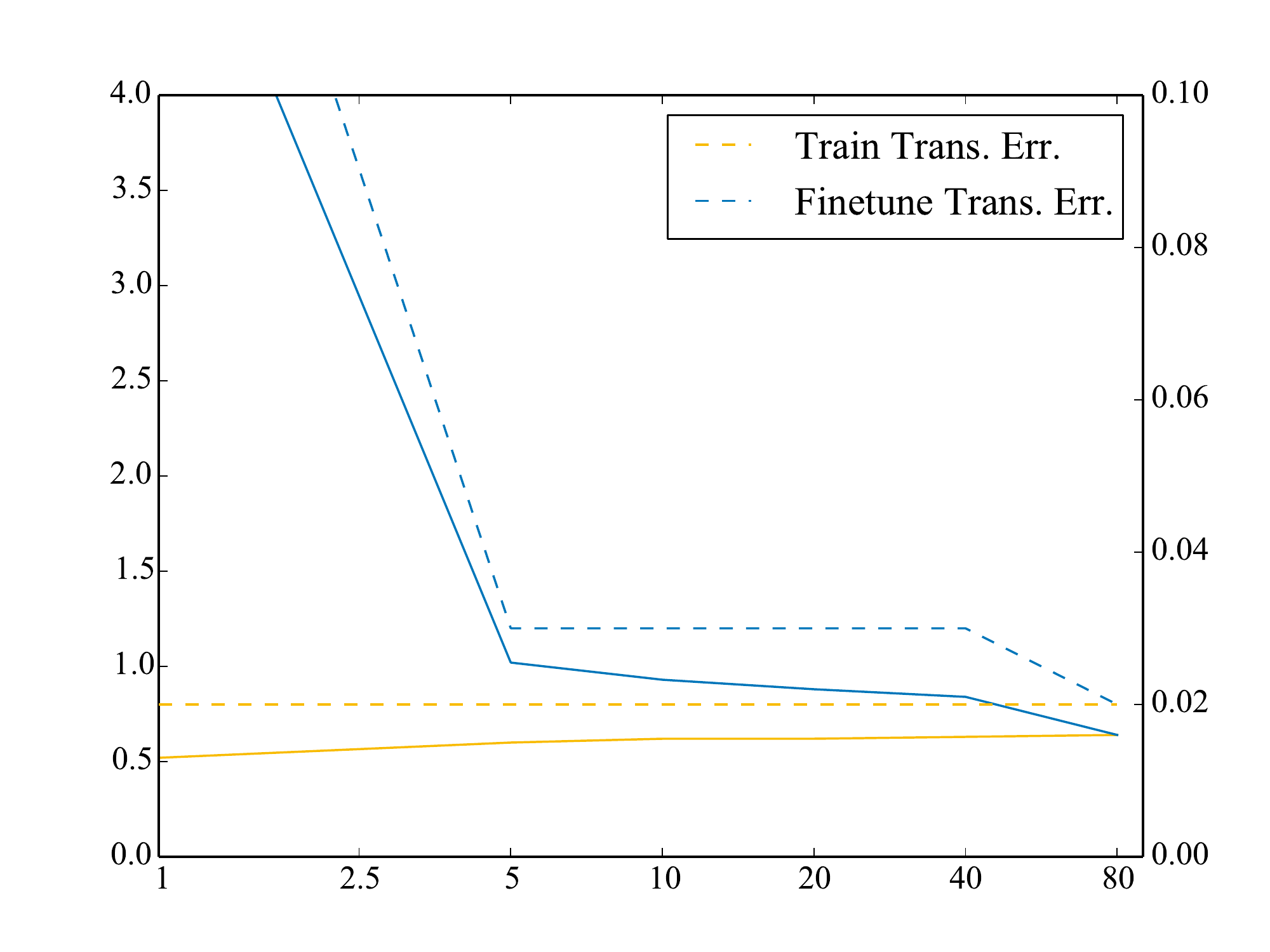}
    \caption{ScanNet results in terms of the median translation and rotation errors (m, $^{\circ}$). We train our model with different numbers of training scenes and finetune tokens on one new testing scene. }
    \label{fig:scannet}
\end{figure}

%% file: table/code_table.tex
\begin{table}[tb]
\begin{tabular}{c|c|c}
Code num.  & Day                                        & Night                                      \\ \hline
1           &  	49.3 / 75.2 / 93.9                      &                 	14.3 / 26.5 / 76.5           \\
25          &   72.2 / 85.4 / 95.4                    &     21.4 / 49.0 / 82.7                                       \\
50          & 	75.6 / 88.8 / 95.5                       & 	33.6 / 59.1 / 83.7                                            \\
100         & 78.0 / 89.0 / 95.2 & 42.9 / 61.2 / 80.6 \\
Pruned (25)  & 76.2 / 88.5 / 95.5 & 37.8 / 62.2 / 87.8                
\end{tabular}
\caption{Performance under different code numbers. We use a voxel of 10 meters to divide the scene on Aachen day $\&$ night dataset and list the code number per transformer block. There are 933 voxels and 6 transformer blocks. It is noted that our network can output good results even with 1 code. For code pruning, there are 25 tokens on average for each transformer block.} \label{tab:code}
\end{table}

%% file: table/block_table.tex
\begin{table}[tb]
\centering
\begin{tabular}{c|c|c}
Block num. & Day                & Night              \\ \hline
1          & 62.1 / 82.8 / 94.4 & 17.3 / 36.7 / 76.5 \\
3          & 68.9 / 85.7 / 94.5 & 20.4 / 40.8 / 79.6 \\
6          & \textbf{78.0} / \textbf{89.0} / \textbf{95.2} & \textbf{42.9} / \textbf{61.2} / \textbf{80.6}
\end{tabular}
\caption{Localization accuracy under different transformer blocks. The performance improves as more blocks are utilized.}\label{tab:block_num}
\end{table}

%% file: tex_file/5_conclusion.tex
\section{Future Work and Limitations}
NeuMap is a simple yet effective localization method with many potentials. Since our framework is end-to-end and fully differentiable, conventional network acceleration technology can be applied for better efficiency, such as code pruning~\cite{molchanov2019importance, liu2018rethinking}, code quantization~\cite{wang2019haq, zhou2017incremental}, and knowledge distillation~\cite{hinton2015distilling, gou2021knowledge}. Besides, the accuracy could also be further boosted by pose loss, commonly used in object pose estimation~\cite{chen2022epro, iwase2021repose}.

A major limitation is the inference speed for large scenes. For high accuracy, a voxel needs to be smaller, increasing the number of voxels to be retrieved. While the coordinate regression is end-to-end, our approach runs the decoder many times for the retrieved voxels and is not necessarily fast (e.g., 5 seconds per image for Aachen). Our future work is better voxel division and retrieval methods for speed up. Another future work is cross-dataset training of NeuMap, which currently degrades performance.


\section{Conclusions}
This paper proposes a novel camera localization method that encodes a scene into a grid of latent codes. Our framework consists of a scene-agnostic neural network and scene-specific latent code. To handle large scenes, we divide a scene into a grid of voxels and assign codes to each. Our method outperforms all the other end-to-end methods by a large margin and achieves similar performance as the feature-matching methods with a much smaller scene representation size. We will share all our code and models.

\paragraph{Acknowledgement} We thank Martin Humenberger and Philippe Weinzaepfel for their codes of ESAC in NAVER LAB datasets. We thank Luwei Yang to evaluate Squeezer under different data sizes. We thank Jiahui Zhang and Jiacheng Chen for proof reading. The research is supported by NSERC Discovery Grants, NSERC Discovery Grants Accelerator Supplements, DND/NSERC Discovery Grant Supplement, and John R. Evans Leaders Fund (JELF).